\documentclass[conference, a4paper, 10pt]{IEEEtran}
\IEEEoverridecommandlockouts
\usepackage{cite}
\usepackage{amsmath,amssymb,amsfonts}
\usepackage{algorithmic}
\usepackage{graphicx}
\usepackage{textcomp}
\usepackage{xcolor}
\def\BibTeX{{\rm B\kern-.05em{\sc i\kern-.025em b}\kern-.08em
    T\kern-.1667em\lower.7ex\hbox{E}\kern-.125emX}}

\usepackage{enumitem}
\usepackage{booktabs}
\usepackage{multirow}
\usepackage[super]{nth}
\usepackage{tikz}
\newcommand*\circled[1]{\tikz[baseline=(char.base)]{
		\node[shape=circle,draw,inner sep=0.2pt] (char) {#1};}}
\newcommand*\circledB[1]{\tikz[baseline=(char.base)]{
            \node[shape=circle,fill,inner sep=0.2pt] (char) {\textcolor{white}{#1}};}}
\usetikzlibrary{tikzmark}
\tikzset{circledColor/.style={circle,draw,inner sep=0.1em,line width=0.04em}}
\usepackage{setspace}

\newcommand{\myCeil}[1]{\left\lceil #1 \right\rceil}

\usepackage{fancyhdr}
\fancypagestyle{firstpage}{
  \fancyhf{}
  \chead{Accepted at the International Joint Conference on Neural Networks (IJCNN), June 30th - July 5th, 2025 in Rome, Italy.}
  \cfoot{\thepage}
}

\begin{document}

\title{Enabling Efficient Processing of Spiking Neural Networks with On-Chip Learning on Commodity Neuromorphic Processors for Edge AI Systems
\vspace{-0.4cm}
}

\author{
\IEEEauthorblockN{Rachmad Vidya Wicaksana Putra, Pasindu Wickramasinghe, Muhammad Shafique}
\IEEEauthorblockA{\textit{eBrain Lab, New York University (NYU) Abu Dhabi, Abu Dhabi, UAE} \\
\{rachmad.putra, pmw6287, muhammad.shafique\}@nyu.edu}
\vspace{-1cm}
}

\maketitle
\pagestyle{plain}
\thispagestyle{firstpage}

\begin{abstract}
The rising demand for energy-efficient edge AI systems (e.g., mobile agents/robots) has increased the interest in neuromorphic computing, since it offers ultra-low power/energy AI computation through spiking neural network (SNN) algorithms on neuromorphic processors. 
However, their efficient implementation strategy has not been comprehensively studied, hence limiting SNN deployments for edge AI systems.  
Toward this, we propose a design methodology to enable efficient SNN processing on commodity neuromorphic processors. 
To do this, we first study the key characteristics of targeted neuromorphic hardware (e.g., memory and compute budgets), and leverage this information to perform compatibility analysis for network selection. 
Afterward, we employ a mapping strategy for efficient SNN implementation on the targeted processor.  
Furthermore, we incorporate an efficient on-chip learning mechanism to update the systems' knowledge for adapting to new input classes and dynamic environments.
The experimental results show that the proposed methodology leads the system to achieve low latency of inference (i.e., less than 50ms for image classification, less than 200ms for real-time object detection in video streaming, and less than 1ms in keyword recognition) and low latency of on-chip learning (i.e., less than 2ms for keyword recognition), while incurring less than 250mW of processing power and less than 15mJ of energy consumption across the respective different applications and scenarios.
These results show the potential of the proposed methodology in enabling efficient edge AI systems for diverse application use-cases.
\end{abstract}


\vspace{-0.2cm}
\section{Introduction}
\label{Sec_Intro}
\vspace{-0.1cm}

\begin{figure}[t]
\centering
\includegraphics[width=0.8\linewidth]{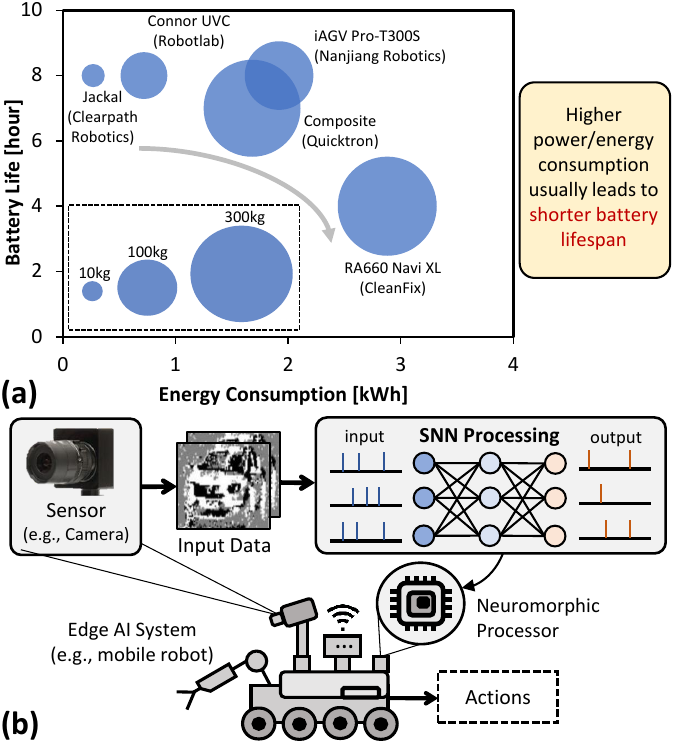}
\vspace{-0.3cm}
\caption{(b) Real-world autonomous systems with higher power/energy consumption usually have heavier mass and shorter battery lifespan; adapted from studies in~\cite{Ref_McNulty_Batteries4AutoMobile_JPS22}. 
(b) Overview of SNN-based computation for edge AI systems, considering an example from a mobile robot use-case.} 
\label{Fig_Problem}
\vspace{-0.7cm}
\end{figure}

In recent years, the demand for employing energy-efficient edge AI systems (e.g., mobile agents/robots and IoT devices) has increased rapidly due to their advantages in improving quality of services (QoS) and human productivity.
However, these systems are typically powered by portable batteries with limited capacity~\cite{Ref_McNulty_Batteries4AutoMobile_JPS22}.
Therefore, they usually suffer from short battery lifespan. 
To address this limitation, employing larger battery capacity may not be a scalable solution, since larger battery capacity means heavier mass, which typically require more power/energy consumption to operate or mobilize the system. 
Current trends also show that, real-world autonomous systems with heavier mass typically consume higher power/energy than the smaller ones, which leads to shorter battery lifespan; see Fig.~\ref{Fig_Problem}(a). 
Therefore, the potential solution is making these systems to employ ultra-low power/energy AI algorithms, thus optimizing the overall systems' energy. 

Toward this, neuromorphic computing (NC) with spiking neural networks (SNN) algorithms has emerged as a potential solution due to its highly sparse spike-driven operations~\cite{Ref_Li_BiC_JPROC24}\cite{Ref_Putra_FSpiNN_TCAD20}; see an illustration in Fig.~\ref{Fig_Problem}(b). 
To maximize the energy-efficiency benefits from NC, SNN processing needs to be performed on specialized neuromorphic hardware processors, which accommodate spike-driven operations~\cite{Ref_Basu_SNNicSurvey_CICC22, Ref_Putra_NeuromorphicAI4Robotics_ICARCV24, Ref_Vogginger_NeuroHW4AIDataCenter_arXiv24}.

Currently, most of the existing neuromorphic processors (e.g., SpiNNAker, NeuroGrid, IBM's TrueNorth, and Intel's Loihi)~\cite{Ref_Basu_SNNicSurvey_CICC22} were developed mainly for research purpose and not commercially available, thereby making it difficult to use them in SNN-based edge AI systems for real-world application use-cases.
Recently, several neuromorphic processors are released and available commercially in the market (such as BrainChip's Akida~\cite{Ref_BrainChip_Akida} and SynSense's DYNAP~\cite{Ref_SynSense_DYNAP}), which can be used for developing real-world SNN-based edge AI systems.
However, their efficient implementation strategy has not been studied, hence limiting the systems from achieving further efficiency gains considering different SNNs and workloads.
Therefore, in this paper, \textbf{the targeted research problem} is: 
\textit{How can we enable efficient execution of SNN models on commodity neuromorphic processors, while achieving good trade-offs between accuracy, memory, and power/energy consumption?}
A solution to this problem may enable energy-efficient SNN deployments for diverse application use-cases at the edge.

\vspace{-0.1cm}
\subsection{State-of-the-Art and Their Limitations}
\label{Sec_Intro_SOTA}
\vspace{-0.1cm}

State-of-the-art works in employing commodity neuromorphic processors for edge AI systems typically focus on the implementation of a specific application use-case, such as system control~\cite{Ref_Dupeyroux_KopohoMAV_ICRA21}~\cite{Ref_Stroobants_KapohoCtrlPID_ICONS22}, tactile sensing~\cite{Ref_Patel_AkidaTactile_AICAS23}, gait analysis~\cite{Ref_Venkatachalam_AkidaGaitAnalysis_ICONS24}, object detection~\cite{Ref_Kadway_AkidaCloudDetect_IJCNN23, Ref_Lenz_AkidaShipDetect_arXiv24, Ref_Silva_AkidaObjDetect_AICAS24}. 
Moreover, the state-of-the-art works have not studied the on-chip learning aspect, which is important for updating the systems' knowledge and adapting to changing environments~\cite{Ref_Putra_SpikeDyn_DAC21}. 
This condition shows that, \textit{a design methodology for enabling efficient SNN execution considering different workloads on commodity neuromorphic processors has not been explored}. 
To show the importance of such a methodology, we conduct a case study in Section~\ref{Sec_Intro_CaseStudy}.

\vspace{-0.1cm}
\subsection{Case Study and Research Challenges}
\label{Sec_Intro_CaseStudy}
\vspace{-0.1cm}

\begin{figure}[t]
\centering
\includegraphics[width=0.95\linewidth]{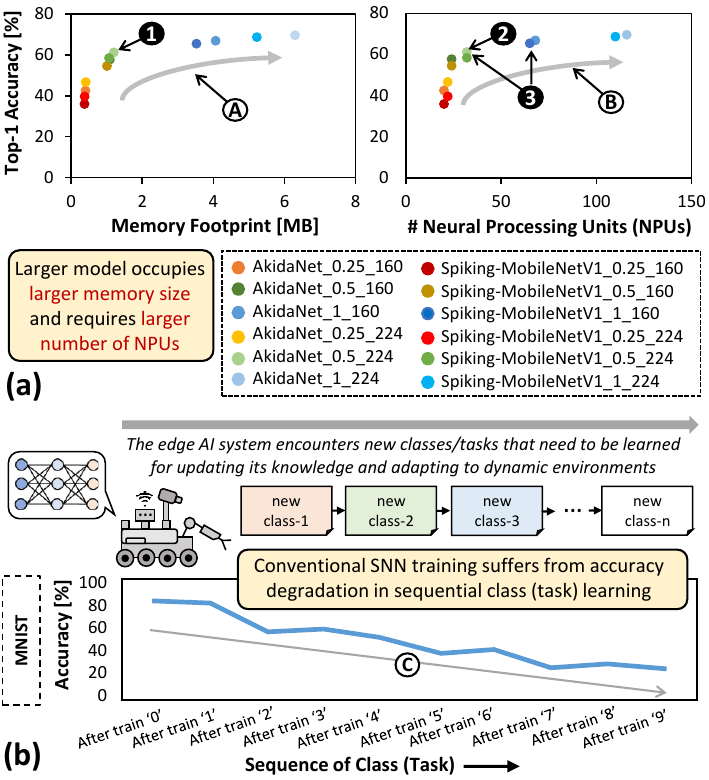}
\vspace{-0.3cm}
\caption{(a) Results of running different SNN models considering different input resolutions on the same commodity neuromorphic processor (i.e., Akida); based on data from~\cite{Ref_BrainChip_MetaTF}. 
Here, each network name denotes ``\textit{Network\_Alpha\_InputResolution}'', where \textit{Alpha} represents the width multiplier that shrinks the network uniformly from the original size.
(b) An edge AI system may encounter new classes at run time that need to be learned for updating its knowledge. 
Conventional SNN training suffers from accuracy degradation in sequential class learning; based on studies in~\cite{Ref_Putra_SpikeDyn_DAC21}.} 
\label{Fig_CaseStudy}
\vspace{-0.5cm}
\end{figure}

We aim to observe the compute and memory requirements for running different SNN models on the same commodity neuromorphic processor.
Here, we consider different SNN models (i.e., AkidaNet and Spiking-MobileNetV1\footnote{AkidaNet is a MobileNetV1-inspired network optimized for deployment on Akida~\cite{Ref_BrainChip_MetaTF}, while Spiking-MobileNetV1 is the converted MobileNetV1 in spiking domain.}) with different input resolutions (i.e., 160x160 and 224x224) that run on the Akida neuromorphic processor\footnote{The details of Akida hardware architecture are provided in Section~\ref{Sec_Back_NeuroChips}, while the details of experimental setup is presented in Section~\ref{Sec_Eval}.}.  
The investigation results are shown in Fig.~\ref{Fig_CaseStudy}(a). 
They show that, running larger SNN model on the neuromorphic processor potentially offer higher accuracy due to higher feature extraction capabilities, but at the cost of larger memory footprint and larger compute resource (i.e., number of Neural Processing Units), and hence higher power/energy consumption; see \circled{A}-\circled{B}. 
Such increased compute, memory, and power/energy requirements can reduce the efficiency gains of SNN-based edge AI systems. 

Edge AI systems may also encounter new class at run time, that need to be learned for updating the systems' knowledge.
Otherwise, these systems can suffer from accuracy degradation when their knowledge becomes obsolete over time~\cite{Ref_Putra_lpSpikeCon_IJCNN22}, as the offline-trained SNN may struggle in training the new classes while preserving the old ones (i.e., previously learned classes)~\cite{Ref_Minhas_NCL_arXiv24}; see \circled{C} in Fig.~\ref{Fig_CaseStudy}(b)

\textit{\textbf{Required:} 
A design methodology that identifies the compatibility of the selected network for the targeted neuromorphic processor, and facilitates on-chip learning mechanism for knowledge updates.}
However, this requirement exposes several research challenges, as described in the following.
\begin{itemize}[leftmargin=*]
    \item The memory and compute costs of the selected network should be efficiently accommodated by the processor.
    \item The selected SNN model needs to be efficiently mapped on the processor to ensure energy-efficient SNN execution.
    \item The system demands an efficient on-chip learning capability to learn new classes, while preserving the knowledge of previously learned ones. 
\end{itemize}

\subsection{Our Novel Contributions}
\label{Sec_Intro_Novelty}

\begin{figure}[t]
\centering
\includegraphics[width=0.78\linewidth]{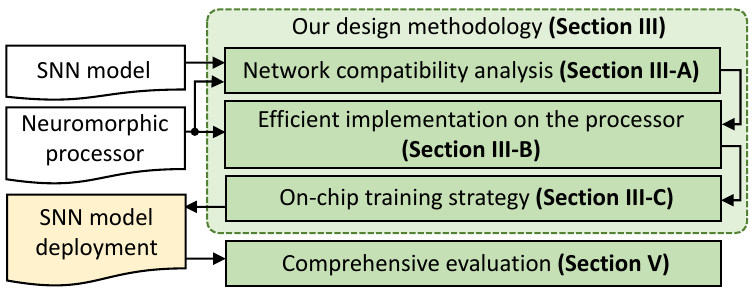}
\vspace{-0.3cm}
\caption{Overview of our novel contributions, highlighted in green.} 
\label{Fig_Novelty}
\vspace{-0.5cm}
\end{figure}

To address the targeted problem and associated challenges, we propose \textit{a novel design methodology that ensures efficient execution of SNNs on commodity neuromorphic processors}, thereby enabling energy-efficient deployments of SNN-based edge AI systems for diverse applications.  
Our novel contributions include the following points (an overview in Fig.~\ref{Fig_Novelty}).
\begin{itemize}[leftmargin=*]
    \item \textbf{A design methodology (Section~\ref{Sec_Method})} that employs several key steps, as summarized below.
    \begin{enumerate}
        \item \textbf{Network compatibility analysis (Section~\ref{Sec_Method_Compatible})}
        that quickly evaluates whether the selected network can be efficiently executed, i.e., by leveraging the characteristics of neuromorphic processor (e.g., memory and compute budgets) and the proposed analytical model.
        \item \textbf{Efficient deployment on the processor (Section~\ref{Sec_Method_Deploy})} by employing an efficient mapping strategy considering good trade-off between hardware costs and efficiency through the processor runtime settings. 
        \item \textbf{On-chip learning strategy (Section~\ref{Sec_Method_OnchipLearn})} that facilitates learning new classes after SNN deployment on the processor through a last layer modification technique. 
    \end{enumerate}
    \item \textbf{Comprehensive evaluation (Section~\ref{Sec_Results})} 
    that covers multiple design metrics, e.g., accuracy, latency, throughput, and power/energy consumption of different SNNs in different applications (i.e., image classification, robject recognition in video streaming, and keyword recognition) and different scenarios (i.e., offline-based and on-chip learning settings).  
\end{itemize}

\textbf{Key Results:}
Our design methodology is evaluated through a real-world edge AI prototype using the Akida neuromorphic processor.
The experimental results show that, our methodology leads the system to achieve low latency of inference (i.e., less than 50ms for image classification, less than 200ms for real-time object detection in video streaming, and less than 1ms in keyword recognition) and low latency of on-chip learning (i.e., less than 2ms for keyword recognition), while consuming less than 250mW of power. 

\section{Background}
\label{Sec_Back}

\subsection{Spiking Neural Networks (SNNs)}
\label{Sec_Back_SNNs}

\begin{figure}[h]
\vspace{-0.3cm}
\centering
\includegraphics[width=0.95\linewidth]{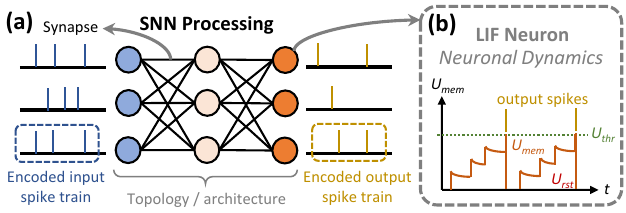}
\vspace{-0.3cm}
\caption{(a) Illustration of an SNN and its components. (b) Overview of the neuronal dynamics of the widely-used spiking neuron model (i.e., LIF).} 
\label{Fig_SNN}
\vspace{-0.2cm}
\end{figure}

\textbf{Overview:}
SNNs are considered the bio-plausible neural network (NN) models~\cite{Ref_Putra_FSpiNN_TCAD20}, since they are modeled after the neural process observed in the human brain, specifically on how neurons utilize spikes for transferring and processing data. 
An SNN model mainly consists of several components: spiking neuron, synapses, network topology/architecture, and neural encoding~\cite{Ref_Putra_QSpiNN_IJCNN21}; see Fig.~\ref{Fig_SNN}(a). 

\textbf{Spiking Neuron:} 
The dynamics of spiking neuron depend on the neuron model, and the widely-used one is the Leaky Integrate-and-Fire (LIF) neuron~\cite{Ref_Putra_ReSpawn_ICCAD21}. 
The neuronal dynamics of LIF is illustrated in Fig.~\ref{Fig_SNN}(b).
Here, $U_{mem}$, $U_{thr}$, and $U_{rst}$ denote the neurons' membrane potential, threshold potential, and reset potential, respectively.
When an incoming spike arrives in the LIF neuron, it triggers the increasing of $U_{mem}$; otherwise, $U_{mem}$ decays. 
If the $U_{mem}$ reaches or surpasses the $U_{thr}$, then an output spike is generated.

\subsection{Neuromorphic Processors}
\label{Sec_Back_NeuroChips}
\vspace{-0.1cm}

\subsubsection{Overview}

The energy efficiency potentials offered by SNNs can be maximized by employing neuromorphic hardware processors~\cite{Ref_Putra_SoftSNN_DAC22}. 
In the literature, several processors have been proposed, and they can be categorized as \textit{research} and \textit{commodity processors}.
Research processors refer to neuromorphic chips that are designed only for research and not commercially available, hence access to these processors is limited. 
Several examples in this category are SpiNNaker, NeuroGrid, IBM's TrueNorth, and Intel's Loihi~\cite{Ref_Basu_SNNicSurvey_CICC22}.
Meanwhile, commodity processors refer to neuromorphic chips that are available commercially, such as BrainChip's Akida~\cite{Ref_BrainChip_Akida} and SynSense's DYNAP-CNN~\cite{Ref_SynSense_DYNAP}. 
\textit{In this work, we consider the Akida processor as it supports on-chip learning for SNN fine-tuning, which is beneficial for adaptive edge AI systems}~\cite{Ref_BrainChip_Akida}.  

\subsubsection{Akida Neuromorphic Processor System-on-Chip}

\begin{figure}[h]
\vspace{-0.3cm}
\centering
\includegraphics[width=0.95\linewidth]{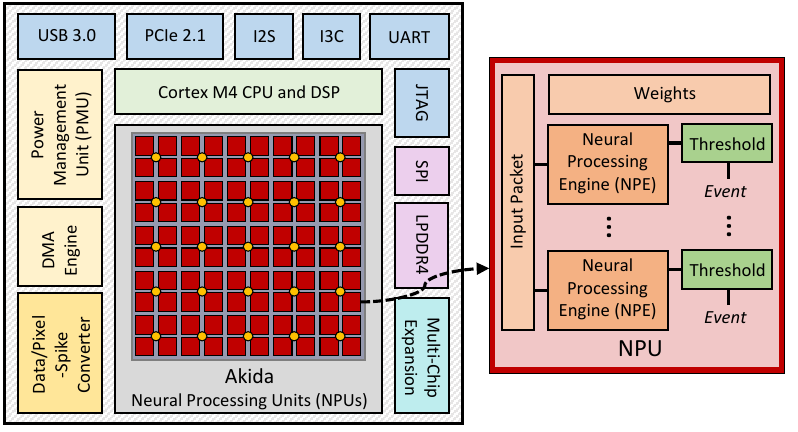}
\vspace{-0.3cm}
\caption{Overview of the Akida architecture; adapted from~\cite{Ref_BrainChip_Akida}\cite{Ref_Demler_Akida_Linley19}.} 
\label{Fig_AkidaArch}
\vspace{-0.4cm}
\end{figure}

Akida Neuromorphic SoC (NSoC) is designed by BrainChip, which aims at accelerating SNN processing for low-power application use-cases~\cite{Ref_BrainChip_Akida}. 
Its commercially available version is the Akida v1.0 (AKD1000)~\cite{Ref_BrainChip_Akida}, which is fabricated using the TSMC's 28nm technology, and it can run at 300MHz clock frequency. 
The overview of Akida NSoC architecture is shown in Fig.~\ref{Fig_AkidaArch}.
It mainly consists of a Cortex-M4 CPU as the SoC host processor and 80 Akida Neural Processing Units (NPUs) as the SNN processors/cores. 
Four NPUs form a single node, hence forming 20 nodes.
Each NPU mainly consists of 8 Neural Processing Engines (NPEs) as the compute units for executing synaptic and neuronal operations (e.g., event-based convolutions), and 100KB SRAM buffer as the local memory for storing weights (40KB) and data spikes (60KB)~\cite{Ref_Demler_Akida_Linley19}. 
Akida NSoC has a direct memory access (DMA) controller, a power management unit (PMU), several data interfaces (i.e., USB 3.0, PCIe 2.1, I2S, I3C, UART, and JTAG), two memory interfaces (i.e., SPI Flash and LPDDR4), and an interface for multi-chip expansion.  
For data encoding, Akida employs the data/pixel-spike converter. 
Furthermore, to facilitate SNN developments and their Akida implementations, BrainChip provides MetaTF framework~\cite{Ref_BrainChip_MetaTF}, which accommodates features like ANN-to-SNN conversion, SNN mapping, and on-chip learning.

\section{Our Design Methodology}
\label{Sec_Method}

\begin{figure*}[t]
\centering
\includegraphics[width=0.83\linewidth]{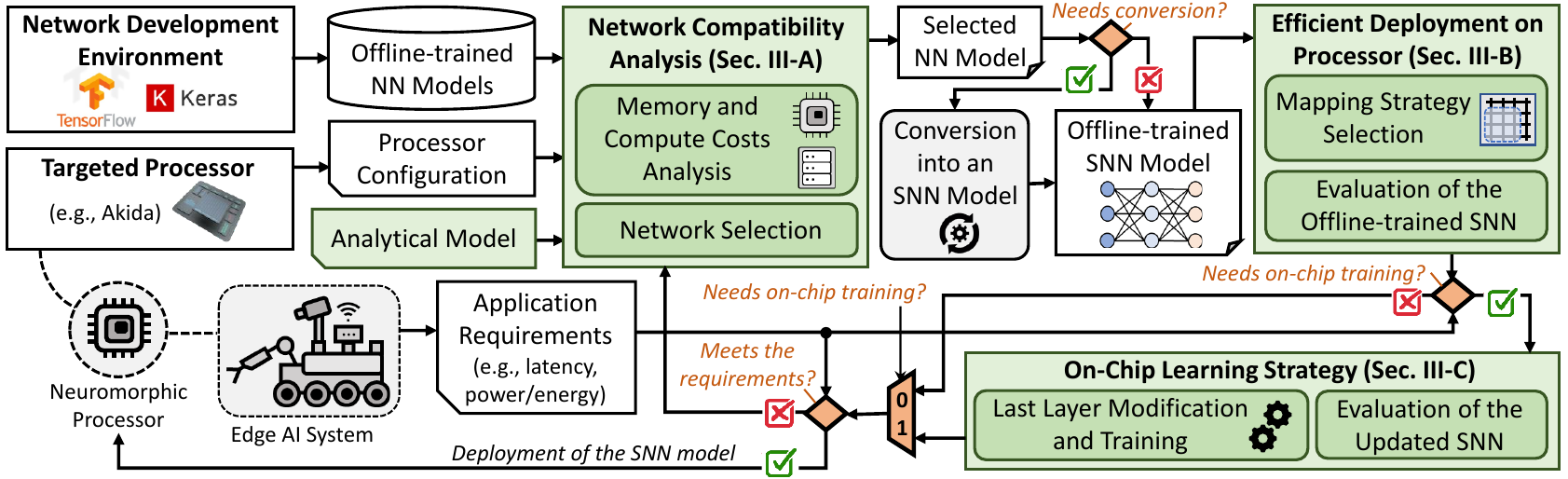}
\vspace{-0.3cm}
\caption{Our proposed design methodology, showing the novel contributions highlighted in green boxes.} 
\label{Fig_ProposedMethod}
\vspace{-0.6cm}
\end{figure*}

We propose a design methodology to address the targeted problem and related challenges, whose key steps are shown in Fig.~\ref{Fig_ProposedMethod}, and discussed in Sections~\ref{Sec_Method_Compatible} until~\ref{Sec_Method_OnchipLearn}. 
For an overview, we describe the flow of our methodology as follows. 
\begin{enumerate}[leftmargin=*]
    \item It starts with the network development using the existing environment (e.g., TensorFlow+Keras) to provide offline-trained NN models.
    \item The offline-trained NN models and the processor configuration are leveraged to perform the network compatibility analysis by using our analytical model for selecting an NN model that meets the memory and compute budgets. 
    \item If the selected NN model is not in spiking domain, we convert the ANN model into an SNN model. Otherwise, the model is already in spiking domain and ready to use.
    \item The SNN model is then deployed on the processor using a specific mapping strategy, and then evaluated with a specific workload (e.g., image classification). 
    \item If a knowledge update is needed, we perform an on-chip learning on the deployed SNN model, and then evaluate it.
    Otherwise, the original SNN is evaluated.
    \item If the evaluated SNN meets the application requirements (e.g., latency and power/energy), then it can be deployed on the processor. Otherwise, we can select a smaller NN model through the network compatibility analysis step. 
    \vspace{-0.1cm}
\end{enumerate}

\subsection{Network Compatibility Analysis}
\label{Sec_Method_Compatible}
\vspace{-0.1cm}

This step aims at \textit{analyzing whether the selected network can be executed efficiently in the targeted neuromorphic processors}. 
Specifically, this refers to the condition where the network can be fully mapped and executed on the processor at one time, and hence no network partitioning and scheduling are required. 
In this manner, costly memory access and data movements can be minimized, as these operations typically dominate the neuromorphic systems' energy~\cite{Ref_Putra_SparkXD_DAC21}. 
This is important because it evaluates the processing requirements in advance before the actual deployment on the hardware, hence guiding the users to better develop and/or select a suitable network to deploy on the targeted processor.

\textbf{Memory and Compute Costs Analysis:}
We identify the characteristics of targeted processor that determine whether the selected network can be fully mapped and executed on the processor at one time, hence avoiding network partitioning and scheduling. 
Specifically, \textit{we investigate the memory and compute budgets, and how they are distributed in the targeted processor}. 
The memory budget represents the maximum size of network and activation data that can be fully mapped at one time, while the compute budget represents the maximum event-based computations that can be executed at one time, hence they are both leveraged for the network compatibility analysis. 
To enable this analysis, we propose \textit{an analytical model} to estimate the memory and compute costs for the given network, while considering the hardware architecture from the processor.
Specifically, it investigates the total number of NPUs (cores) required to fully map and execute the given network and data (i.e., denoted as $N_{NPU\_tot}$).

\textbf{Poposed Analytical Model:}
$N_{NPU\_tot}$ is defined as a total number of NPUs required across different layers of the given network; see Eq.~\ref{Eq_NPUtot}. 
Here, $L$ represents the number of layers in the network, and $N^l_{NPU\_mem}$ represents the number of NPUs required for storing network parameters and data in layer-$l$.
\begin{equation}
    \begin{split}
    N_{NPU\_tot} = \sum^{L}_{l=1} N^l_{NPU\_mem} 
    \end{split}
    \label{Eq_NPUtot}
\end{equation}
\noindent We observe that, SNN parameters and data in the same layer may have different sizes, hence requiring different numbers of NPUs. 
To properly allocate hardware resources for such a condition, we select the bigger number of NPUs to ensure sufficient memory resource, whose function can be can stated as Eq.~\ref{Eq_NPUl}. 
Here, $N^l_{NPU\_net}$ and $N^l_{NPU\_dat}$ denote the number of NPUs for network parameters and data in layer-$l$, respectively. 
\begin{equation}
    \begin{split}
    N^l_{NPU\_mem} = \text{max}(N^l_{NPU\_net}, N^l_{NPU\_dat}) 
    \end{split}
    \label{Eq_NPUl}
\end{equation}
We can obtain $N^l_{NPU\_net}$ and $N^l_{NPU\_dat}$ using Eq.~\ref{Eq_NPUnetdat}. 
Here, $M^l_{net}$ and $M^l_{dat}$ denote the size of network parameters and activation data in layer-$l$, respectively.
Meanwhile, $B_{net}$ and $B_{dat}$ denote the local memory (buffer) size in each NPU for network parameters and data, respectively.
\begin{equation}
    \begin{split}
    N^l_{NPU\_net} = \myCeil{\frac{M^l_{net}}{B_{net}}} \;\; \text{and} \;\; 
    N^l_{NPU\_dat} = \myCeil{\frac{M^l_{dat}}{B_{dat}}}
    \end{split}
    \label{Eq_NPUnetdat}
\end{equation}
Furthermore, $M^l_{net}$ can be obtained by leveraging the number of parameters (i.e., weights $N_w$ and bias $N_b$) with their bit precision ($bit_{par}$) in layer-$l$; see Eq.~\ref{Eq_MemNet}. 
Meanwhile, $M^l_{dat}$ can be obtained by leveraging the number of feature maps with their bit precision ($bit_{dat}$) in layer-$l$; see Eq.~\ref{Eq_MemDat}.
Note, $H^l$, $W^l$, and $C^l$ denote the feature maps' dimension in layer-$l$ for height, width, and channel, respectively.
\begin{equation}
    \begin{split}
    M^l_{net} & = (N^l_w + N^l_b) \cdot bit^l_{par}
    \end{split}
    \label{Eq_MemNet}
\end{equation}
\begin{equation}
    \begin{split}
    M^l_{dat} = (H^l \cdot W^l \cdot C^l) \cdot bit^l_{dat}
    \end{split}
    \label{Eq_MemDat}
\end{equation}

\textbf{Network Selection:}
We use the proposed analytical model to identify the network models that can be efficiently executed in the targeted processor, i.e., by selecting the network models whose memory and compute costs ($N_{NPU\_tot}$) are less than the memory and compute budgets from the processor ($N_{NPU\_proc}$), while offering high accuracy.

\subsection{Efficient Deployment on the Processor}
\label{Sec_Method_Deploy}

This step aims to \textit{enable efficient deployment of the selected network on the processor}.
It requires a mapping strategy that leads the SNN processing to meet the application requirements (e.g., latency and power).
In practice, the possible strategies also depend on the availability of related application programming interface (API) of the chip's implementation framework.

\textbf{Mapping Strategy:} 
In this work, we employ a mapping strategy from the Akida's MetaTF framework~\cite{Ref_BrainChip_MetaTF} that \textit{maximizes hardware resources for holding network parameters and data with minimum passes (i.e., sequential processing), thus providing a trade-off between performance and efficiency}.
This strategy aims at mapping the entire SNN model and data in the NPU local memories, while employing a sequential processing for synaptic and neuronal operations in each layer, minimizing parallel NPU processing. 
Therefore, the optimization objective is to minimize the number of NPUs for executing synaptic and neuronal operations for each layer ($N^l_{NPU\_exe}$); see Eq.~\ref{Eq_Optim}.
\begin{equation}
    \begin{split}
    Objective: \text{minimize} (N^l_{NPU\_exe})
    \end{split}
    \label{Eq_Optim}
\end{equation}
The cost function for hardware mapping ($C$) is defined as the total NPU allocation for storing network parameters and data ($N_{NPU\_mem}$) and executing synaptic and neuronal operations (optimized $N_{NPU\_exe}$) across all layers ($L$). 
To allocate resources for such a condition, we select the bigger number of NPUs for facilitating each layer processing; see Eq.~\ref{Eq_Cost}.
\begin{equation}
    \begin{split}
    C = \sum^{L}_{l=1} \{ \text{max}(N^l_{NPU\_mem}, N^l_{NPU\_exe}) \}
    \end{split}
    \label{Eq_Cost}
\end{equation}
Consequently, the hardware utilization ($U$) can be determined through the ratio between the mapping cost $C$ and the total number of NPUs in the processor $N_{NPU\_proc}$; see Eq.~\ref{Eq_Util}.
\begin{equation}
    \begin{split}
    U = \frac{C}{N_{NPU\_proc}} \cdot 100\% 
    \end{split}
    \label{Eq_Util}
\end{equation} 

\textbf{Network Evaluation:}
After mapping the selected SNN on the processor, we evaluate its performance and efficiency to observe if the selected SNN meets the given requirements.
If the systems need to update their knowledge, then this SNN model needs to be updated through an on-chip learning. 

\subsection{On-Chip Learning Strategy}
\label{Sec_Method_OnchipLearn}

It aims at \textit{learning new classes on-chip, thereby enabling an efficient fine-tuning for the existing SNN model}. 
Here, possible on-chip learning strategies for neuromorphic processors mainly also depend on the availability of related API from their development framework. 

\textbf{On-chip Learning:}
For a case study, we use the available on-chip learning strategy from the Akida's MetaTF framework~\cite{Ref_BrainChip_MetaTF}.
To enable the on-chip learning, we need to fulfill the following learning constraints for the last network layer, since this last layer is the only part that will be trained in on-chip learning: (1) it must be a fully connected type, (2) it must have binary weights, and (3) it must receive binary inputs.
We fulfill these requirements through the following steps.
\begin{itemize}[leftmargin=*]
    \item We replace the last layer with a new layer that meets the learning constraints/characteristics.
    \item The new layer should accommodate classifying both the old and new classes. 
    Hence, multiple neurons for each old class are used to provide spaces for learning new classes on-chip. 
    \item We perform a few-shot learning on-chip with a few samples for each new class using the Akida's built-in algorithm. 
    The new classes are associated with specific neurons in last layer. 
\end{itemize}

\textbf{Network Evaluation:}
After mapping the updated SNN on the processor, we evaluate its performance and efficiency to observe if the updated SNN meets the given requirements.

\section{Evaluation Methodology}
\label{Sec_Eval}

\begin{figure}[h]
\vspace{-0.2cm}
\centering
\includegraphics[width=0.98\linewidth]{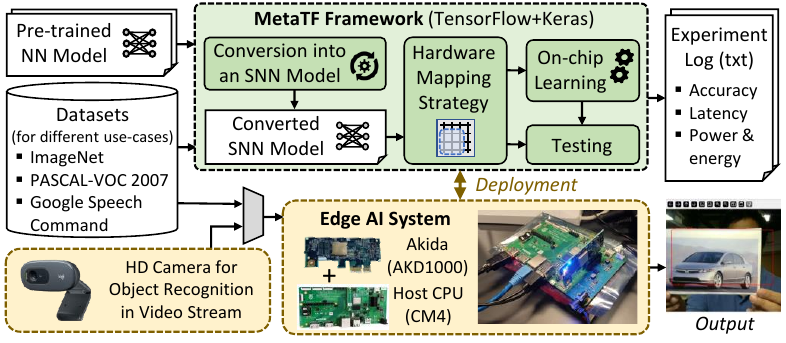}
\vspace{-0.4cm}
\caption{Experimental setup and tools flow. 
} 
\label{Fig_ExpSetup}
\vspace{-0.2cm}
\end{figure}

To evaluate our proposed design methodology, we employ the experimental setup and tools flow presented in Fig.~\ref{Fig_ExpSetup}. 

\textbf{Software Development Part:}
We employ MetaTF framework~\cite{Ref_BrainChip_MetaTF}, which is based on TensorFlow and Keras libraries, to convert the pre-trained NN model into an SNN model. 
Afterward, this SNN model is mapped on the neuromorphic processor and executed accordingly.
For the on-chip learning, the SNN model is modified to facilitate learning new classes; as discussed in Section~\ref{Sec_Method_OnchipLearn}. 
In experiments, we record results like accuracy, latency, and power/energy consumption.  

\textbf{Hardware Development Part:}
We develop a real-world edge AI system, comprising neuromorphic processor and host CPU.  
For the neuromorphic processor, we employ a single Akida NSoC (Akida v1.0 AKD1000)~\cite{Ref_BrainChip_Akida}.
Meanwhile, for the host CPU, we employ an ARM Cortex-A72 through the Raspberry Pi Compute Module 4 (CM4), which runs the Ubuntu 22.04 OS. 
These host and neuromorphic processors are connected through the PCIe interface, thereby providing a high-speed serial computer expansion bus. 

\begin{figure*}[t]
\centering
\includegraphics[width=\linewidth]{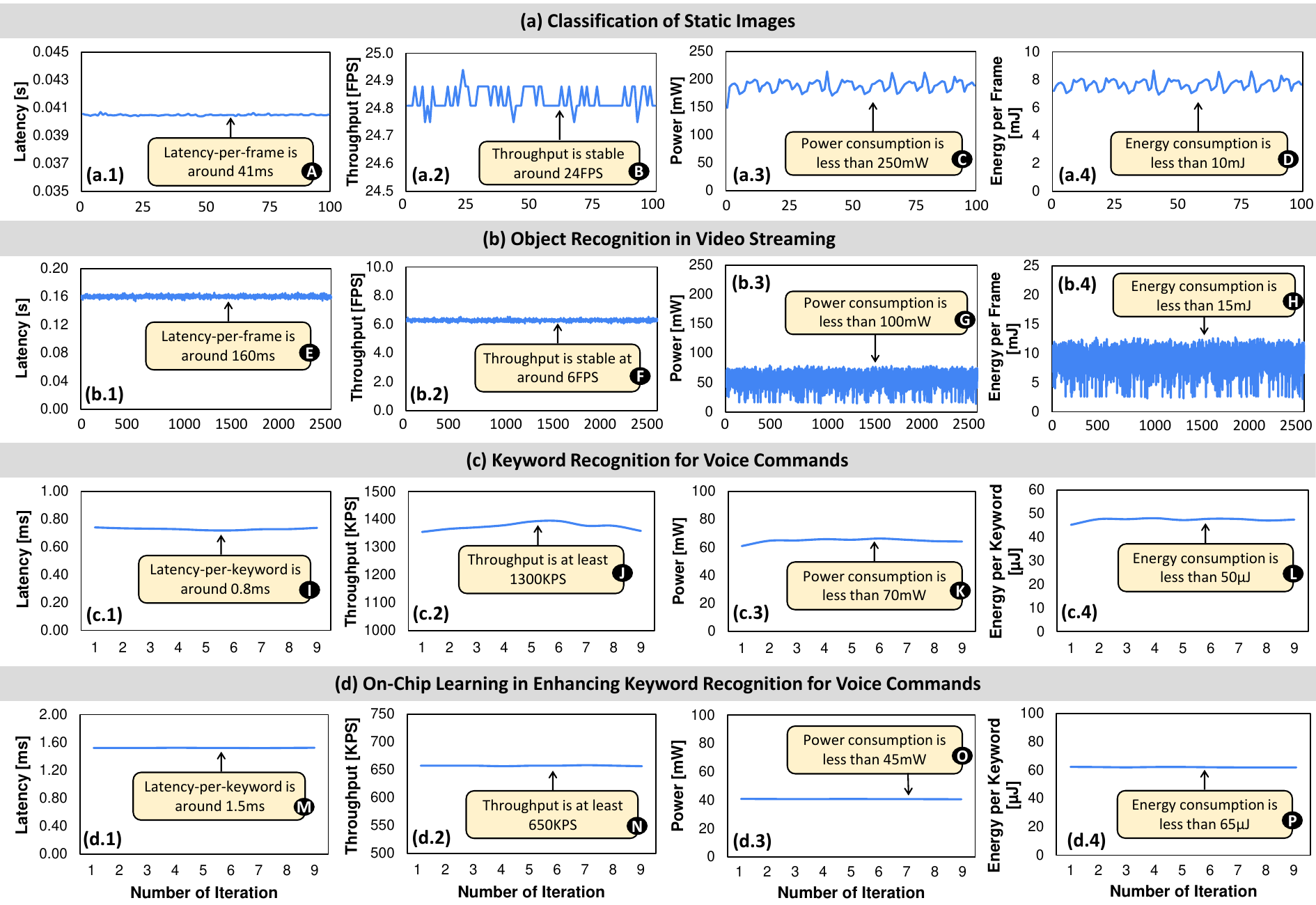}
\vspace{-0.7cm}
\caption{Experimental results of running SNN models on Akida for (a) classification of static images, (b) real-time object detection in video stream, (c) keyword recognition for voice commands, and (d) on-chip learning in enhancing keyword recognition for voice commands; encompassing the latency, throughput, as well as power and energy consumption. 
Note, FPS denotes the frame-per-second, while KPS denotes the keyword-per-second.} 
\label{Fig_Results_PerfEfficient}
\vspace{-0.4cm}
\end{figure*}

\textbf{Application Use-Cases:}
For showing the generality of our design methodology, we consider three different applications. 
\begin{enumerate}[leftmargin=*]
    \item \textit{Classification of Static Images:} 
    It considers the ImageNet dataset~\cite{Ref_Deng_ImageNet_CVPR09}. 
    Its application requirements include the maximum 50ms latency and 250mW power consumption. 
    \item \textit{Real-time Object Recognition in Video Streaming:}
    It considers the PASCAL-VOC 2007 dataset~\cite{PascalVOC2007}. 
    We perform real-time object detection in video streaming using a complete edge AI system utilizing a Logitech C270 HD WebCam; see Fig.~\ref{Fig_ExpSetup}.
    Its application requirements include the maximum 200ms latency and 250mW power consumption. 
    \item \textit{Keyword Recognition:}
    It uses the Google Speech Command dataset~\cite{Ref_Warden_GoogleSpeechCommand_arXiv18}. 
    Its application requirements include the maximum 5ms latency and 250mW power consumption.
\end{enumerate}

\section{Results and Discussion}
\label{Sec_Results}

\subsection{Classification of Static Images}
\label{Sec_Results_ImageClass}

\textbf{Network Selection:}
Our network compatibility analysis (discussed in Section~\ref{Sec_Method_Compatible}) leads to the selection of AkidaNet\_0.5\_224 from many possible network models due to the following reasons.
\begin{itemize}[leftmargin=*]
    \item The network size meets the memory budget of an Akida chip (i.e., 8MB), as shown by~\circledB{1} in Fig.~\ref{Fig_CaseStudy}(a).
    \item The number of NPUs required for a complete computation of the network meets the NPU budget of an Akida chip (i.e., 80 NPUs), as shown by~\circledB{2} in Fig.~\ref{Fig_CaseStudy}(a).
    \item The network achieves higher accuracy as compared to other network models, but slightly lower than AkidaNet\_1\_160 and Spiking-MobileNetV1\_1\_160, as shown by~\circledB{3} in Fig.~\ref{Fig_CaseStudy}(a). 
    We select AkidaNet\_0.5\_224 as it can handle higher input resolution 224x224, which is beneficial for systems with high resolution sensors. 
\end{itemize}

\textbf{Accuracy:} 
Here, we perform inference by presenting 10 images from the ImageNet to the system over 100 iterations. 
The experimental results show that, running AkidaNet\_0.5\_224 on the Akida processor achieves 80\% accuracy. 
This high accuracy comes from an effective training process in ANN domain that employs the accurate backpropagation technique, and an effective conversion technique that accurately translates the trained ANN components and parameters into representative SNN components and parameters.

\textbf{Latency, Throughput, Power and Energy Consumption:}
The results for latency and throughput are shown in Fig.~\ref{Fig_Results_PerfEfficient}(a.1) and Fig.~\ref{Fig_Results_PerfEfficient}(a.2), respectively.
Latency is stable around 41ms across 100 iterations of experiments (see \circledB{A}), which leads to 24FPS throughput (see \circledB{B}), thereby meeting the design requirement of maximum 50ms latency. 
Such low latency and high throughput mainly come from the selected mapping strategy, which considers maximizing hardware resources to fully map the entire SNN model on the Akida's NPU fabrics, hence avoiding the time-consuming execution of network partitions.
Meanwhile, the results for power and energy consumption are presented in Fig.~\ref{Fig_Results_PerfEfficient}(a.3) and Fig.~\ref{Fig_Results_PerfEfficient}(a.4), respectively.
Overall, power consumption is about 215mW (see \circledB{C}), and energy consumption is about 9mJ (see \circledB{D}), thereby meeting the design requirement of maximum 250mW power. 
Such low power and low energy consumption come from the sparse spike-driven computation, and the selected mapping strategy that optimizes data movements through efficient NPU allocation, and hence minimizing power consumption for the respective operations. 

\subsection{Real-Time Object Detection in Video Streaming}
\label{Sec_Results_ObjDet}

\textbf{Network Selection:}
Our compatibility analysis (from Section~\ref{Sec_Method_Compatible}) leads to the selection of Spiking-YOLOv2~\cite{Ref_BrainChip_MetaTF} due to the following reasons.
\begin{itemize}[leftmargin=*]
    \item The network size (i.e., $\sim$3MB) meets the memory budget of Akida chip (i.e., 8MB).
    \item The number of NPUs required for a complete computation of the network (i.e., 71 NPUs) meets the NPU budget of an Akida chip (i.e., 80 NPUs).
\end{itemize}

\begin{figure}[h]
\vspace{-0.2cm}
\centering
\includegraphics[width=0.95\linewidth]{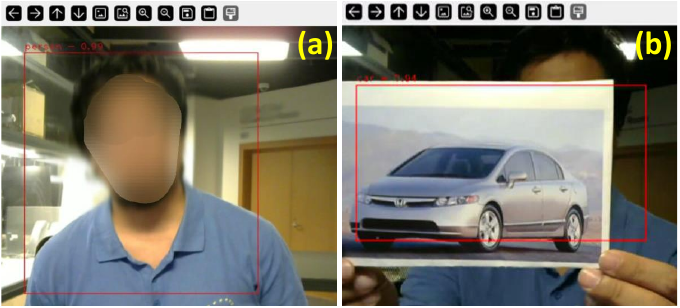}
\vspace{-0.3cm}
\caption{Experimental results of the edge AI system prototyping based on the Akida neuromorphic processor for object detection in video streaming: (a) person detection, and (b) car detection.} 
\label{Fig_Res_ObjDetDemo}
\vspace{-0.2cm}
\end{figure}

\textbf{Accuracy:} 
We perform inference using with 2500 iterations of object presentation (i.e., person and car). 
Screenshots of the real-time object detection in video streaming are presented in Fig.~\ref{Fig_Res_ObjDetDemo}. 
The experimental results show that, running Spiking-YOLOv2 on the Akida processor can achieve 94.44\% accuracy for detecting the presented objects.
This high accuracy is due to the training process that utilizes accurate backpropagation in ANN domain, and the conversion process that accurately translates the ANN model into a representative SNN model.

\textbf{Latency, Throughput, Power and Energy Consumption:}
The experimental results for latency and throughput are shown in Fig.~\ref{Fig_Results_PerfEfficient}(b.1) and Fig.~\ref{Fig_Results_PerfEfficient}(b.2), respectively.
Processing latency is around 160ms across 2500 iterations of experiments (see \circledB{E}), which leads to 6FPS throughput (see \circledB{F}), and thereby meeting the design requirement of maximum 200ms latency. 
These results show that, the system achieves relatively low latency and high throughput for real-time object detection in video streaming. 
It is due to the selected mapping strategy, which fully maps the entire SNN model on the NPU fabrics, which avoids the time-consuming execution of network partitions. 
Meanwhile, the experimental results for power and energy consumption are shown in Fig.~\ref{Fig_Results_PerfEfficient}(b.3) and Fig.~\ref{Fig_Results_PerfEfficient}(b.4), respectively.
Power consumption is $\sim$78mW (see \circledB{G}), and energy consumption is $\sim$13mJ (see \circledB{H}), thereby meeting the design requirement of maximum 250mW power. 
Such low power and low energy consumption come from the sparse spike-driven computation, and the selected mapping strategy that optimizes data movements through judicious NPU allocation, hence minimizing the power consumption. 

\subsection{Keyword Recognition for Voice Commands}
\label{Sec_Results_KWS}

\textbf{Network Selection:}
Our compatibility analysis (from Section~\ref{Sec_Method_Compatible}) leads to the selection of Spiking Depthwise Separable Convolutional Neural Network (Spiking-DSCNN)~\cite{Ref_BrainChip_MetaTF} due to the following reasons.
\begin{itemize}[leftmargin=*]
    \item The network size (i.e., 23KB) meets the memory budget of Akida chip (i.e., 8MB).
    \item The number of NPUs required for a complete computation of the network (i.e., 5 NPUs) meets the NPU budget of an Akida chip (i.e., 80 NPUs).
\end{itemize}

\textbf{Accuracy:} 
Here, we perform inference by presenting 1000 keywords from the Google Speech Command dataset to the system over 9 iterations. 
The experimental results show that, running Spiking-DSCNN on the Akida processor can achieve 91.73\% accuracy.
This high accuracy is due to an effective ANN training and its accurate ANN-to-SNN conversion. 

\textbf{Latency, Throughput, Power and Energy Consumption:}
The experimental results for processing latency and throughput are presented in Fig.~\ref{Fig_Results_PerfEfficient}(c.1) and Fig.~\ref{Fig_Results_PerfEfficient}(c.2), respectively.
Here, processing latency is stable around 0.72ms across 9 iterations of experiments (see \circledB{I}), which leads to more than 1300KPS (keyword-per-second) throughput (see \circledB{J}), thereby meeting the design requirement of maximum 5ms latency. 
Such low latency and high throughput mainly come from the small size of Spiking-DCNN with 23KB, which makes the entire network easy to map on the NPU fabrics.
Consequently, this avoids the time-consuming execution of network partitions, while incurring small data movements and operations.  
Meanwhile, the experimental results for power and energy consumption are presented in Fig.~\ref{Fig_Results_PerfEfficient}(c.3) and Fig.~\ref{Fig_Results_PerfEfficient}(c.4), respectively.
Power consumption is about 68mW (see \circledB{K}), and energy consumption is about 49$\mu$J (see \circledB{L}), thereby meeting the design requirement of maximum 250mW power. 
These low power and low energy consumption mainly come from the small size of Spiking-DCNN which makes the entire network model can be efficiently executed on the NPU fabrics. 

\begin{table*}[t]
\caption{Summary of comparison between our neuromorphic platform (Akida) with existing conventional AI solutions for object detection using YOLOv2; based on our results and data from state-of-the-art~\cite{liu2022yolov2, nakahara2018lightweight, yan2021fpga}.}
\vspace{-0.1cm}
\scriptsize
\centering
\begin{tabular}{c|c|c|c|c|ccc|c|}
\cline{2-9}
\multicolumn{1}{l|}{} & \multicolumn{1}{c|}{\textbf{Desktop CPU}} & \multicolumn{1}{c|}{\textbf{Desktop GPU}} & \multicolumn{1}{c|}{\textbf{Embedded CPU}} & \multicolumn{1}{c|}{\textbf{Embedded GPU}} & \multicolumn{3}{c|}{\textbf{FPGA}} & \multicolumn{1}{c|}{\multirow{2}{*}{\textbf{\begin{tabular}[c]{@{}c@{}}Our Akida\\ Neuromorphic\\ Platform \end{tabular}}}} \\ \cline{2-8}
\multicolumn{1}{l|}{}                                                                                       & \multicolumn{1}{c|}{\textbf{\begin{tabular}[c]{@{}c@{}}Intel \\ i7-6700HQ\end{tabular}}} & \multicolumn{1}{c|}{\textbf{\begin{tabular}[c]{@{}c@{}}Nvidia \\ GTX 960M\end{tabular}}} & \multicolumn{1}{c|}{\textbf{\begin{tabular}[c]{@{}c@{}}ARM \\ Cortex-A57\end{tabular}}} & \multicolumn{1}{c|}{\textbf{\begin{tabular}[c]{@{}c@{}}Nvidia\\ Jetson TX2\end{tabular}}} & \multicolumn{1}{c|}{\textbf{ZedBoard}} & \multicolumn{1}{c|}{\textbf{\begin{tabular}[c]{@{}c@{}}ZCU102\end{tabular}}} & \multicolumn{1}{c|}{\textbf{\begin{tabular}[c]{@{}c@{}}Virtex-7\\ XC7V690t\end{tabular}}} & \multicolumn{1}{c|}{} \\ \hline
\multicolumn{1}{|c|}{\textbf{\begin{tabular}[c]{@{}c@{}}Performance {[}FPS{]}\end{tabular}}}        & 78.2 & 219.7 & 0.23 & 7.8 & 1.02 & 40.81 & 302.3 & \textbf{6} \\ 
\multicolumn{1}{|c|}{\textbf{Power {[}W{]}}} & 29.88 & 46.67 & 4 & 5.8 & 1.2 & 4.5 & 11.35 & \textbf{0.078} \\ 
\multicolumn{1}{|c|}{\textbf{\begin{tabular}[c]{@{}c@{}}Efficiency {[}FPS/W{]}\end{tabular}}} & 2.62 & 4.71 & 0.06 & 1.34 & 0.85 & 9.06 & 26.63 & \textbf{76.92} \\ \cline{1-9}
\end{tabular}
\label{Tab_Comparison}
\vspace{-0.4cm}
\end{table*}
\setlength{\textfloatsep}{2pt}
\subsection{On-Chip Learning for Knowledge Updates}
\label{Sec_Results_OnchipLearn}

We select the keyword recognition application and employ the pre-trained Spiking-DSCNN. 
From the the Google Speech Command dataset, we use 32 keywords for the offline training and 3 new keywords (i.e., ‘backward’, ‘follow’, and ‘forward’) as new classes for on-chip learning. 

\textbf{Accuracy:}
We first perform on-chip learning on the Akida for 3 new keywords, and each one is trained using 160 samples. 
Then, we perform inference using 6 samples for ‘backward’, 7 samples for ‘follow’, and 6 samples for ‘forward’ over 9 iterations. 
The experimental results show that, running Spiking-DSCNN on the Akida can achieve 94.74\% accuracy.
This high accuracy is due to the effective few-shot learning algorithm provided by the Akida's MetaTF framework.

\textbf{Latency, Throughput, Power and Energy Consumption:}
The experimental results for latency and throughput for on-chip learning with the Akida are presented in Fig.~\ref{Fig_Results_PerfEfficient}(d.1) and Fig.~\ref{Fig_Results_PerfEfficient}(d.2), respectively.
Latency is stable around 1.5ms across 9 iterations of experiments (see \circledB{M}), which leads to more than 650KPS throughput (see \circledB{N}). 
Such low latency and high throughput come from the efficient few-shot learning that utilizes relatively small number of samples for new classes.
The same reason also leads the on-chip learning to incur low power and energy consumption.
Power consumption is about 41mW (see \circledB{O}), and energy consumption is about 62$\mu$J (see \circledB{P}), as shown in Fig.~\ref{Fig_Results_PerfEfficient}(d.3) and Fig.~\ref{Fig_Results_PerfEfficient}(d.4), respectively.

\vspace{-0.1cm}
\subsection{Further Discussion}
\label{Sec_Results_Discuss}
\vspace{-0.1cm}

It is important to compare neuromorphic-based solutions against the state-of-the-art ANN-based solutions, which typically employ conventional hardware platforms, such as CPUs, GPUs, and specialized accelerators (e.g., FPGA or ASIC). 
To ensure a fair comparison, we select object recognition as the application and YOLOv2 as the network, while considering performance efficiency (FPS/W) as the comparison metric. 
Summary of the comparison is provided in Table~\ref{Tab_Comparison}, and it clearly shows that our Akida-based neuromorphic solution achieves the highest performance efficiency.
This is due to the sparse spike-driven computation that is fully exploited by neuromorphic processor, thus delivering highly power/energy-efficient SNN processing.
Moreover, our Akida-based neuromorphic solution also offers an on-chip learning capability, which gives it further advantages over the other solutions.
This comparison highlights the immense potentials of neuromorphic computing for enabling efficient edge AI systems.

\vspace{-0.1cm}
\section{Conclusion}
\label{Sec_Conclude}
\vspace{-0.1cm}

We propose a novel design methodology to enable efficient SNN processing on commodity neuromorphic processors. 
It is evaluated using a real-world edge AI system implementation with the Akida processor. 
The experimental results demonstrate that, our methodology leads the system to achieve high performance and high energy efficiency across different applications. 
It achieves low latency of inference (i.e., less than 50ms for image classification, less than 200ms for real-time object detection in video streaming, and less than 1ms for keyword recognition) and low latency of on-chip learning (i.e., less than 2ms for keyword recognition), while consuming less than 250mW of power. 
In this manner, our design methodology potentially enables ultra-low power/energy design of edge AI systems for diverse application use-cases.

\vspace{-0.1cm}
\section*{Acknowledgment}
\vspace{-0.1cm}
This work was partially supported by the NYUAD Center for Artificial Intelligence and Robotics (CAIR), funded by Tamkeen under the NYUAD Research Institute Award CG010.

\begin{spacing}{0.98}
\bibliographystyle{IEEEtran}
\bibliography{bibliography}
\end{spacing}
\end{document}